\documentclass[sigconf, screen, authorversion]{acmart}    

\usepackage{booktabs} 
\usepackage{algorithm}
\usepackage{algorithmic}
\usepackage{amsmath}
\usepackage{amssymb}
\usepackage{stmaryrd}
\usepackage{float}
\usepackage{threeparttable}

\usepackage{tikz}

\usepackage{colortbl}

\usepackage{pgfplots}
\usepgfplotslibrary{groupplots}
\pgfplotsset{compat=1.12}

\usepackage[normalem]{ulem}
\useunder{\uline}{\ul}{}

\usepackage{graphicx}

\begin{document}
\title{A Novel Online Stacked Ensemble for Multi-Label Stream Classification}

\author{Alican B\"uy\"uk\c{c}ak{\i}r}
\orcid{0000-0002-6987-6671}
\affiliation{%
  \institution{Bilkent Information Retrieval Group \\ Computer Engineering Department}{Bilkent University}
  \postcode{06800}
}
\email{alicanbuyukcakir@bilkent.edu.tr}

\author{Hamed Bonab}
\affiliation{%
  \institution{College of Information and Computer Sciences}{University of Massachusetts Amherst}
  \postcode{01003}
  }
\email{bonab@cs.umass.edu}

\author{Fazli Can}
\orcid{0000-0003-0016-4278}
\affiliation{%
  \institution{Bilkent Information Retrieval Group \\ Computer Engineering Department}{Bilkent University}
  \postcode{06800}
}
\email{canf@cs.bilkent.edu.tr}


\renewcommand{\shortauthors}{B\"uy\"uk\c{c}ak{\i}r et al.}

\begin{abstract}
As data streams become more prevalent, the necessity for online algorithms that mine this transient and dynamic data becomes clearer. Multi-label data stream classification is a supervised learning problem where each instance in the data stream is classified into one or more pre-defined sets of labels. Many methods have been proposed to tackle this problem, including but not limited to ensemble-based methods. Some of these ensemble-based methods are specifically designed to work with certain multi-label base classifiers; some others employ online bagging schemes to build their ensembles. In this study, we introduce a novel online and dynamically-weighted stacked ensemble for multi-label classification, called GOOWE-ML, that utilizes spatial modeling to assign optimal weights to its component classifiers. Our model can be used with any existing incremental multi-label classification algorithm as its base classifier. We conduct experiments with 4 GOOWE-ML-based multi-label ensembles and 7 baseline models on 7 real-world datasets from diverse areas of interest. Our experiments show that GOOWE-ML ensembles yield consistently better results in terms of predictive performance in almost all of the datasets, with respect to the other prominent ensemble models.
\end{abstract}

%
%

\copyrightyear{2018} 
\acmYear{2018} 
\setcopyright{acmcopyright}
\acmConference[CIKM '18]{The 27th ACM International Conference on Information and Knowledge Management}{October 22--26, 2018}{Torino, Italy}
\acmBooktitle{The 27th ACM International Conference on Information and Knowledge Management (CIKM '18), October 22--26, 2018, Torino, Italy}
\acmPrice{15.00}

\ccsdesc[500]{Information systems~Data stream mining}
\ccsdesc[500]{Computing methodologies~Ensemble methods}
\ccsdesc[300]{Computing methodologies~Online learning settings}

\keywords{Multi-label; data stream; supervised learning; classification; ensemble learning; bagging; online learning}

\maketitle

\section{Introduction}

The traditional supervised learning task is single-label, i.e. a data instance is classified into one label $\lambda$ among a disjoint set of labels $\mathcal{L}$. However, this may not be the case for some real-world data. For instance, \textit{The Big Lebowski} can simultaneously be classified as a \textit{crime}, \textit{comedy}, and \textit{cult} movie. In such settings where an instance can be classified into a subset of labels, $L^{\ast} \subseteq \mathcal{L}$, the learning paradigm is called Multi-label Learning (MLL).

As of 2017, it is estimated that around 4.9 billion connected devices are generating data and this number is expected to rise to 25 billion by 2020 \cite{CACMStaff:2017:BD:3098997.3079064}. With such a rate of increase in the number of data in the form of streams, it becomes more and more important to extract meaningful information from seemingly chaotic data. Some of these data streams are multi-label, which led to MLL algorithms that can cope with streaming settings (i.e. with time and memory constraints, as well as changes in the distribution of the data over time) being developed.

MLL algorithms have drawn considerable attention over the last decades by accomplishing strong results in diverse areas including bioinformatics \cite{clare2001knowledge, barutcuoglu2006hierarchical}, text classification \cite{lewis2004rcv1} and image scene classification \cite{boutell2004scene}. A considerable number of MLL algorithms resort to ensemble methods to increase their predictive performances \cite{read2009classifier, read2008pruned, qu2009mining, wang2017weighted}. However, these methods have usually employed online bagging schemes for ensemble construction (where some of them utilized change detection mechanisms \cite{bifet2007adwin} as an upgrade to these bagging-based ensembles). To the best of our knowledge, there are very few stacked ensembles for multi-label stream classification, and most of the stacked ensembles in the literature are designed for and can only work with specific types of MLL algorithms. In this paper, we propose a novel stacked ensemble that is agnostic of the type of multi-label classifier that is used within the ensemble.

\begin{table*}[!h]
\centering
\vspace{-0.3cm}
\caption{Symbols and Notation for Multi-Label Stream Classification}
\vspace{-0.4cm}
\label{table-of-symbols}
\begin{tabular}{ p{2cm} p{11cm}}
\textbf{Symbol}           & \textbf{Meaning}           \\
\hline
M                & Number of attributes in a data instance \\
L                & Number of labels in the labelset of a data instance \\
N                & Number of instances in the data stream \\
\hline
$\mathcal{X}$              & Input attribute space. $\mathcal{X} = \mathbb{R}^M$\\
$x$              & A data instance.  $x = <x_1, x_2, .., x_i, .., x_M> \in \mathcal{X}$ \\
$\mathcal{L}$    & Set of all possible labels. $\mathcal{L} = \{ \lambda_1, \lambda_2, .., \lambda_L \}$ \\
$y$              & Label relevance vector. $y = <y_1, y_2, .., y_j, .., y_{L}> = \{0,1\}^L$\\
$\hat{y}$          & Predicted relevance vector. $\hat{y} = h(x) = <\hat{y}_1, \hat{y}_2, .., \hat{y}_j, .., \hat{y}_L> = [0,1]^L$\\ 
$d_t = (x^t, y^t)$     & The data point that arrives at time t\\
$\mathcal{D}$  & Possibly infinite data stream. $\mathcal{D} = d_0, d_1, .., d_t, .., d_N$\\
\hline
\end{tabular}
\vspace{-0.3cm}
\end{table*}

The main contributions of this paper are as follows: We (1) introduce a batch-incremental, online stacked ensemble for multi-label stream classification, GOOWE-ML, that can work with any incremental multi-label classifier as its component classifiers; (2) construct an $|\mathcal{L}|$ dimensional space to represent the relevance scores of classifiers of the ensemble, and utilize this construction to assign optimum weights to the model's component classifiers; (3) conduct experiments on 7 real-world datasets to compare GOOWE-ML with 7 state-of-the-art ensemble methods; (4) apply statistical tests to show that our model outperforms the state-of-the-art multi-label ensemble models. Additionally, we discuss how and why some multi-label classifiers yield poor Hamming Scores while performing considerably well on the rest of the performance metrics (e.g. accuracy, F1 Score). All in all, we argue that GOOWE-ML is well-suited for the multi-label stream classification task, and it is a valuable addition to the present day models.

The rest of the paper is organized as follows: Section 2 gives preliminaries on multi-label stream classification. Section 3 introduces the most widely used multi-label algorithms and ensemble techniques in the literature. In Section 4, our ensemble, GOOWE-ML, is described with the theory behind it. After mentioning the experimental setup, evaluation metrics and datasets in Section 5, the results are presented and discussed in Section 6. Lastly, the paper is concluded with insights and possible future work.

\section{Preliminaries}

MLL is considered to be a hard task by nature, as the output space increases exponentially with the number of labels, since there are $2^L$ possible outcomes of classification for the labelset of size $L$. The high dimensionality of the label space causes increased computational cost, execution time and memory consumption. \textit{Multi-label stream classification} (MLSC \cite{wang2017weighted}) is the version of this task that takes place on data streams.

Data stream $\mathcal{D}$ is the set of data that has a temporal dimension and is possibly infinite. $\mathcal{D} = d_0, d_1, .., d_t, .., d_N$ where $d_t$ is the data point in time t and $d_N$ is the lastly seen data point in the data stream. The knowledge of lastly seen data point $d_N$ is not known a priori, and it is only there to indicate the end of the processed data instances for evaluation purposes. Each data point $d_t$ is of form $d_t = (x, y)$ where $x$ is a data instance and $y$ is its labelset (label relevance vector). The data instance $x$ is a vector represented as $x = <x_1, x_2, .., x_i, .., x_M>$, and each $x_i \in \mathcal{X}$. The labelset $y$ is a vector represented as $y = <y_1, y_2, .., y_j, .., y_{L}>$, and each $y_j \in \{0, 1\}$. Here, $y_j = 1$ means the $j$th label is relevant, and 0 otherwise. A prediction (hypothesis) of a multi-label classifier is $\hat{y} = h(x)$ that is of form $\hat{y} = <\hat{y}_1, \hat{y}_2, .., \hat{y}_j, .., \hat{y}_L>$ and $\hat{y} \in [0,1]^L$ meaning that the prediction vector consists of relevance probabilities (relevance scores) for each label. For the final decision of classification and evaluation, the prediction vector is sent to \textit{de-fuzzification}, typically done by thresholding the relevance scores \cite{sorower2010literature}.

\section{Related Work}

Comprehensive reviews on multi-label learning can be found at \cite{sorower2010literature, zhang2014review, gibaja2014review}, on ensemble learning for data streams at \cite{Krawczyk2017132, Gomes:2017:SEL:3071073.3054925} and ensemble of multi-label classifiers at \cite{moyano2018review}. In this paper, we discuss the state-of-the-art multi-label methods, and focus on how these methods are used in ensemble learners for data streams.

\subsection{Multi-label Methods}

As a widely accepted taxonomy in the field of MLL, there are two general methods \cite{zhang2014review} of tackling a multi-label classification problem: 

\subsubsection{Problem Transformation}

In \textit{Problem Transformation}, the multi-label problem is transformed into more well-understood and simpler problems. 

The most widely used Problem Transformation method is the Binary Relevance (BR) \cite{tsoumakas2011rakel} where the multi-label problem is transformed into $|\mathcal{L}|$ distinct binary classification problems. After the transformation is applied to the dataset, any off-the-shelf binary classification algorithm can be utilized to get individual outputs corresponding to each binary problem. It scales linearly with respect to the number of labels, which makes it an efficient choice for practical purposes. However, it has been discussed \cite{zhang2014review, read2009classifier} that BR inherently fails to capture label-wise interdependencies. 

To capture dependencies among labels and overcome this weakness of BR, some other BR-based methods are developed; most notably Classifier Chains (CC) \cite{read2009classifier} where BR classifiers are randomly permuted and linked in a chain-like manner in which each BR classifier yields its output to its connected neighbor classifier as an attribute. It is claimed that this helps the classifiers to capture the label dependencies, as each classifier in the chain learns not only the data itself, but also the label associations of every previous classifier in the chain. 

Another common method of Problem Transformation is Label Powerset (LP) \cite{tsoumakas2011rakel} method where each possible subset of labels is treated as a single label to translate the initial problem into single-label classification task with a bigger set of labels (hence, having multi-class problem of size $2^{|\mathcal{L}|}$). Pruned Sets (PS) \cite{read2008pruned} is an LP-based technique where the instances with infrequent label sets are pruned from the dataset. This allows only the instances with the most important subsets of labels to be considered for classification. Afterwards, the pruned instances are recycled back into an auxiliary dataset for another phase of classification; but for every subset of their relevant labels, instead of their initial relevant labels.

\subsubsection{Algorithm Adaptation}

In \textit{Algorithm Adaptation}, existing algorithms are modified to be compatible with the multi-label setting.

In ML-KNN \cite{zhang2007ml}, the k-Nearest Neighbor algorithm is modified by counting the number of relevant labels for each neighboring instance to acquire posterior relevance probabilities for labels. 

\begin{figure*}[!ht]
\includegraphics[width=0.82\linewidth]{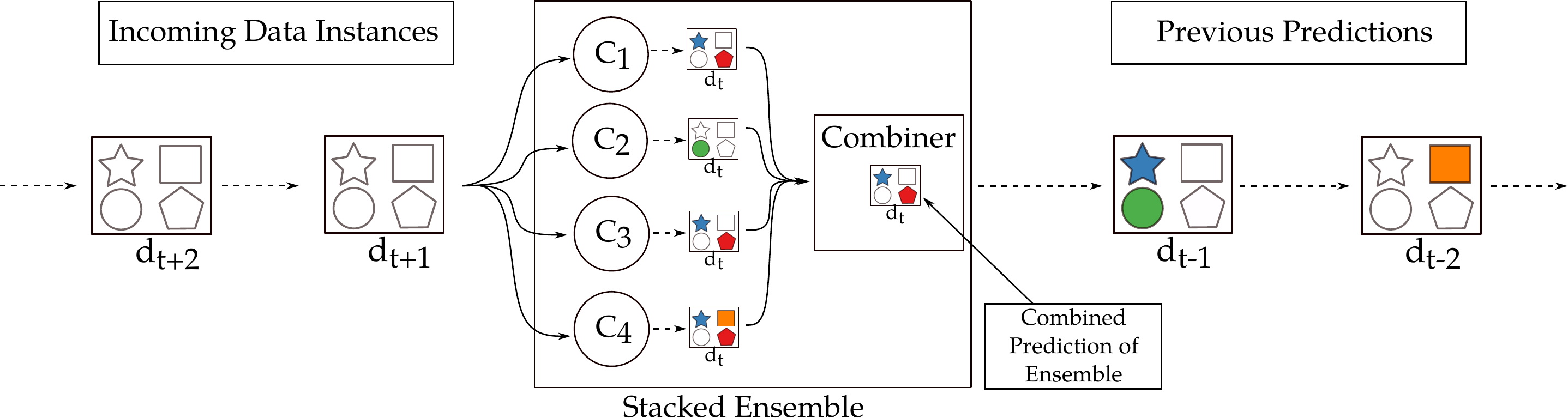}
\vspace{-0.3cm} \caption{A stacked multi-label ensemble for stream classification. For each data instance, associated labels are shown with the geometric shapes ($\square$, $\bigcirc$, and so on). A shape is colored if that label is relevant. Component classifiers ($C_1, C_2, C_3, C_4$) generate their own predictions, and these predictions are combined by the combiner algorithm of the ensemble.}
\label{fig:ml-ensemble}
\end{figure*}

In ML-DT \cite{clare2001knowledge}, the split criterion of C4.5 decision trees is modified by introducing the concept of multi-label entropy. In streaming environments, however, Hoeffding Trees are the common choice for decision trees. Hoeffding Trees \cite{domingos2000mining} are incremental decision trees that have a theoretical guarantee that their output will become asymptotically identical to that of a regular decision tree as more and more data instances arrive. Modifying the split criterion of Hoeffding Trees for multi-label entropy, Multi-label Hoeffding Trees \cite{read2012scalable} are developed. More recently, a novel decision tree based method, iSOUP-Trees (incremental Structured Output Prediction Tree) \cite{osojnik2017isoup} are proposed where adaptive perceptrons are placed in the leaves of incremental trees and the perceptrons' weights are used in producing a prediction that is a linear combination of the input's attributes.

In a nutshell, ``in Problem Transformation, data is modified to make it suitable for algorithms; whereas in Algorithm Adaptation, algorithms are modified to make them suitable for data" \cite{zhang2014review}.

\subsection{Ensembles in MLL and MLSC}



One of the most commonly used ensemble methods is Bagging where each classifier in an ensemble is trained with a bootstrap sample (a data sample that has the same size with the dataset, but each data point is randomly drawn with replacement). This assumes that the whole dataset is available, which is not the case for data stream environments. However, observing that the probability of having K many of a certain data point in a bootstrap sample is approximately \texttt{Poisson(1)} for big datasets, each incoming data instance in a data stream can be weighted proportional to \texttt{Poisson(1)} distribution to mimic bootstrapping in an online setting \cite{oza2005onlinebb}. This is called Online Bagging, or OzaBagging, and it has been widely used in MLSC. In fact, the phrase \textit{`Ensemble of'} in the field usually means that it is the OzaBagged version of the base classifier that is mentioned. EBR \cite{read2009classifier}, ECC \cite{read2009classifier}, EPS \cite{read2008pruned} and EBRT \cite{osojnik2017isoup} (Ensembles of BR, CC, PS and iSOUP Regression Trees respectively) are examples of this convention.

Additionally, it is common for the ensembles that use OzaBagging to also use a concept change detection mechanism called ADWIN (Adaptive Windowing) \cite{bifet2007adwin}. ADWIN keeps a variable-length window of the most recent items in the data stream to detect diversions from the average of the values in the window. Therefore, whenever ADWIN detects a change, the worst classifier in the OzaBag is reset. This is called ADWIN Bagging \cite{read2012scalable}.

To the best of our knowledge, stacked ensemble models in the field of MLSC are very rare. A general scheme for a stacked ensemble for MLSC is given in Figure \ref{fig:ml-ensemble}. Predictions of the component classifiers of an ensemble should be combined by a function (a meta-classifier) which will generate the final prediction of the ensemble. One can use either raw confidence scores of the labels for each instance, or predictions for each label and their counts (majority voting scheme) as the contributions of each component. How to optimally combine the contributions of each classifier is still a question in MLSC. Stacked ensembles that are proposed in the field are as follows: 

SWMEC \cite{wang2017weighted} is a weighted ensemble that is designed for ML-KNN as its base classifier. Its weight adjustment scheme utilizes distances in ML-KNN to obtain a confidence coefficient. IBR (Improved BR) \cite{qu2009mining} employs a feature extension mechanism in which the outputs of a BR classifier is firstly weighted by the accuracy of that classifier, and then added as a new feature to the data instance. New BR classifiers are trained from the data with extended feature spaces. The functionalities of these models involve algorithm-specific properties and therefore cannot be extended to any other base classifier. Such models are constrained by the success of their base classifiers. In \cite{wang2012mining}, the authors followed an unorthodox approach and created a \textit{label-based} ensemble instead of a chunk-based one, which tackled the class imbalance problem that exists in the multi-label datasets as well as concept drifts. Recently, in ML-AMRules \cite{sousa2018multi}, multi-label classification task is interpreted as a rule learning task and the rule learners are combined in an ensemble that uses online bagging (called ML-Random Rules).

All in all, ensemble models in MLSC are not explored thoroughly. Base multi-label classifiers are either combined with Online Bagging or ADWIN Bagging, or with stacked combination schemes that depends on the type of the base classifier. There is a lack of online ensembles in the field that can work with any type of multi-label base classifier which also involve a smart combination scheme. GOOWE-ML addresses this inadequacy.

\section{GOOWE-ML}


\begin{table*}[!h]
\centering
\caption{Additional Symbols and Notation for GOOWE-ML}
\vspace{-0.3cm}
\label{table-of-symbols-goowe}
\begin{tabular}{ l l }
\textbf{Symbol}           & \textbf{Meaning}           \\
\hline
$K$                & Number of component classifiers in the ensemble, i.e. ensemble size \\
$n$                & Number of data points in the instance window I \\
$h$                & Maximum capacity of a data chunk DC \\
\hline
$C_k$            & $k$th component classifier in the ensemble. $ 1 \leq k \leq K $ \\
$\xi$            & Ensemble of classifiers. $\xi = \{ C_1, C_2, .., C_k, .., C_K \}$ \\
$w$              & Weight vector for the ensemble $\xi$. $w = <W_1, W_2, .., W_k, .., W_K>$ \\
$s_k^i$          & Relevance scores for the $k$th classifier for $i$th instance in the ensemble. $s^i_k = <S^i_{k1}, S^i_{k2}, .., S^i_{kj}, .., S^i_{kL}>$ \\
$S$              & Relevance scores matrix. Each relevance score $s_{kj}$ is an element in this matrix. $S \in \mathbb{R}^{K \times L}$ \\
$I$              & Instance window of size $n$, having latest $n$ data instances. $I = d_1, d_2, .., d_n$ \\
$DC$             & Data chunk that consists of the latest $h$ data points. $DC = d_1, d_2, .., d_h$ \\
\hline
\end{tabular}
\end{table*}

We propose GOOWE-ML (\textbf{G}eometrically \textbf{O}ptimum \textbf{O}nline \textbf{W}eighted \textbf{E}nsemble for \textbf{M}ulti-\textbf{L}abel Classification): a batch-incremental (chunk-based) and dynamically-weighted online ensemble that can be used with any incremental multi-label learner that yields confidence outputs for predicting relevant labels for an incoming data instance.

Let the multi-label classifiers in the ensemble be $\{ C_1, C_2, \dots, C_K \}$. For each incoming data instance, each classifier $C_k$ generates relevance score vector $s_k$, which consists of the relevance scores of each label for that instance, i.e. $s_k = < S_{k1}, S_{k2}, \dots, S_{kL} >$. The relevance score vectors of classifiers for each instance is stored in the rows of matrix $S$, which will be used to populate elements of the matrix $A$ and the vector $d$ (see Eqn.\ref{eqnfora} and \ref{eqnford}, and Alg.\ref{alg:goowe-weight}:\ref{code:AeqnLine}-\ref{code:deqnLine}).

\subsection{Ensemble Maintenance}

Let $\xi$ denote the ensemble that is initially empty. A new classifier is trained at each incoming data chunk, as well as the existing ones (if any). The ensemble grows as the new classifiers from incoming data chunks are introduced, until the maximum ensemble size is reached (i.e. ensemble is full). Then, the newly trained classifier replaces one of the old classifiers in the ensemble. This replacement is often times done by removing the temporally oldest component or the most poorly performed component with respect to some metric \cite{kuncheva2004classifier}. In GOOWE-ML, this replacement is done by re-weighting the component classifiers and removing the component with the lowest weight (Alg \ref{alg:goowe}:\ref{code:replaceOldStart}-\ref{code:replaceOld}). Analogous model management systems are employed in both ensembles for single-label classification such as Accuracy Weighted Ensemble (AWE) \cite{wang2003awe} and Accuracy Updated Ensemble (AUE2) \cite{brzezinski2014aue2}; and for multi-label classification such as SWMEC \cite{wang2017weighted}.

Having a fixed number of base classifiers in the ensemble prevents the model to swell in terms of memory usage. Also, training new classifiers from each data chunk allows the ensemble to notice new trends in the distribution of the data, and thus, be more robust against concept drifts.

In addition to fixed-sized data chunks, GOOWE-ML also uses a sliding window for stream evaluation purposes, which consists of the most recently seen $n$ instances. Size of the instance window can be smaller than the size of each data chunk, i.e. $n \leq h$, so that higher resolution can be obtained for the prequential evaluation. Prequential evaluation is discussed in more detail in Experimental Setup section. 

\subsection{Weight Assignment and Update}

\begin{figure}[!ht]
\includegraphics[width=0.8\columnwidth]{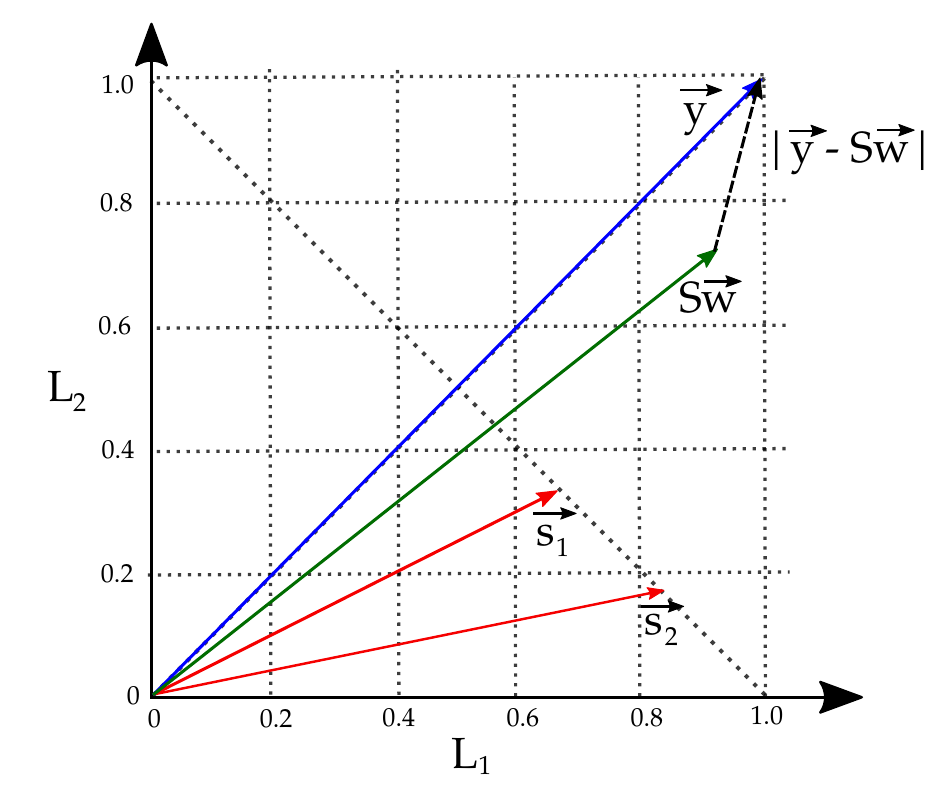}
\vspace{-0.3cm} \caption{Transformation into label space in GOOWE-ML. Relevance scores of the components (red): $S_1 = <0.65,0.35>$ and $S_2 = <0.82,0.18>$. The optimal vector $\vec{y}$ (blue): $y = <1,1>$, generated from the ground truth. Weighted prediction of the ensemble: $S\vec{w}$ (green). The distance between $\vec{y}$ and $S\vec{w}$ is minimized.}
\label{labelspace}
\end{figure}

In our geometric framework, we represent the relevance scores $s_k$ of each component classifier in our ensemble as vectors in an $L$-dimensional space. Previously, Tai \& Lin \cite{tai2012plst} used a similar approach, which they called Principal Label-Space Transformation, to interpret the existing multi-label algorithms in a geometrical setting and reduce the high dimensionality of the multi-label data. Bonab \& Can \cite{bonab2016theoretical} adapted an analogous setting to investigate the optimal ensemble size for single-label data stream classification. In GOOWE-ML, this spatial modeling is used to assign optimal weights for component classifiers in the ensemble.

Geometrically, an intuitive explanation of our spatial modeling and weighting scheme is shown in Figure \ref{labelspace} for the 2-dimensional case, i.e. when $L=2$. After representing relevance scores in the label space, GOOWE-ML minimizes the Euclidean distance between the combined relevance vector, $\hat{y}$, and the ideal vector that represents the ground truth, $y$, in the label space. Analogously, Wu \& Crestani \cite{wucrestani2015} utilized this approach in the field of Data Fusion to optimally combine the query results, and Bonab \& Can \cite{bonab2018goowe} in the field of single-label data stream classification, both with successful results. This is equivalent to the following \textit{linear least squares problem}.

\begin{equation}
    \min_{\vec{w}} || \vec{y} - S\vec{w} ||_2^2
\end{equation}

Here, $S$ is the relevance scores matrix, $w$ is the weight vector which is to be determined, and $y$ is the vector representing the ground truth for a given data point. In other words, our objective function to be minimized is the following:

\begin{equation}
\\
  f(W_1, W_2, .., W_K) = \sum_{i=1}^n \sum_{j=1}^L \bigg( \sum_{k=1}^K ( W_k S^i_{kj} - y^i_j ) \bigg)^2
\end{equation}

Taking a partial derivative of $W$ and setting the gradient to zero, i.e. $\nabla f = 0$, we get:

\begin{equation}
\label{transformeqn}
  \sum_{k=1}^K W_k \bigg( \sum_{i=1}^n \sum_{j=1}^L  S^i_{qj} S^i_{kj} \bigg) =  \sum_{i=1}^n \sum_{j=1}^L  y^i_j S^i_{qj}
\end{equation}

Equation \ref{transformeqn} is of the form $Aw = d$ where A is a square matrix of size $K \times K$ with elements:

\begin{equation}
\label{eqnfora}
  a^i_{qk} = \sum_{i=1}^n \sum_{j=1}^L S^i_{qj} S^i_{kj} \quad \quad (1 \leq q,k \leq K)
\end{equation}

and $d$ is the remainder vector of size $K$ with elements:

\begin{equation}
\label{eqnford}
  d^i_q = \sum_{i=1}^n \sum_{j=1}^L y^i_j S^i_{qj} \quad \quad (1 \leq q \leq K)
\end{equation}

Therefore, solving the equation $Aw = d$, for $w$, gives us the optimally adjusted weight vector. The weight vector $w$ is updated at the end of each data chunk, where the components in the ensemble are trained from the instances in the data chunk, as well. Also, notice that this update operation resembles the Batch Gradient Descent in a way that $w$ is updated at the end of each batch, having trained from the instances in the batch. However, unlike Batch Gradient Descent, this weight update scheme does not take steps towards better weights iteratively, but rather finds the optimal weights directly after solving the linear system $Aw=d$. As a consequence, the updated weights do not depend on the previous values in $w$, they only depend on the performance of the components on the latest chunk. This allows the ensemble to capture sudden changes in the distribution of the data. 

\subsection{Multi-label Prediction}

The ensemble's prediction for the $i$th example, $\hat{y}^i$, is the weighted sum of its components' relevance scores, $s_k$. 

\begin{equation}
\label{weightedvoting}
\hat{y}^i_j (\xi) = \sum_{k=1}^K w_k S^i_{kj} \quad \quad   (1 \leq j \leq L)
\end{equation}

Here, each relevance score, $S^i_{kj}$, is normalized beforehand into the range of $[0,1]$ by the following normalization:

\begin{equation}
\label{normalization}
S^i_{kj} \leftarrow \frac{S^i_{kj}}{\sum_{j=1}^L S^i_{kj}} \quad \quad (1 \leq j \leq L)
\end{equation}

After normalization, the relevance scores sum up to 1. The final prediction of the classifier is obtained by thresholding the relevance scores by $(1 / L)$, which is the expected prior relevance probability of a label of a data instance.

\begin{equation}
\label{thresholding}
    \hat{y}^i_j \leftarrow \begin{cases} 
      1, & \text{if } \hat{y}^i_j > \dfrac{1}{L} \\
      0, & otherwise 
   \end{cases}  \quad \quad 1 \leq j \leq L
\end{equation}

These three operations (Weighted Voting (Eqn.\ref{weightedvoting}), Normalization (Eqn.\ref{normalization}) and Thresholding (Eqn.\ref{thresholding})) are done consecutively and can be considered as one atomic operation in the algorithm, shown as \textit{predict()} in the pseudocode (see Alg.\ref{alg:goowe}:\ref{code:test}).

\begin{algorithm}
    \caption{GOOWE-ML: Geometrically Optimum Online Weighted Ensemble for Multi-Label Classification}
    \label{alg:goowe}
    \begin{algorithmic}[1]
        \REQUIRE $\mathcal{D}$: data stream, $DC$: latest data chunk, $K$: maximum number of classifiers in the ensemble, $C$: a multi-label classifier in the ensemble,
        \ENSURE $\xi$: ensemble of weighted classifiers, $\hat{y}$: multi-label prediction of the ensemble as combined score vector.
        \STATE $\xi \leftarrow \emptyset$;
        \STATE $A, d \leftarrow null, null$
        \WHILE{$\mathcal{D}$ has more instances}
            \STATE $d_i \leftarrow $ current data instance  
            \STATE $\hat{y} \leftarrow $predict$(d_i, \xi )$ \hfill \COMMENT{Eqn.\ref{weightedvoting}, \ref{normalization} and \ref{thresholding}} \label{code:test}
            
            \IF{$DC$ is full}
                \STATE $C_{in} \leftarrow $ new component classifier built on $DC$; \label{code:replaceOldStart}
                \IF{$\xi$ has $K$ classifiers}
                    \STATE $A', d' \leftarrow $TrainOptimumWeights$(DC, \xi, null, null)$ \label{code:trainoptimumweights}
                    \STATE $w \leftarrow $ solve$(A'w = d')$;
                    \STATE $C_{out} \leftarrow $ classifier $C_k$ with minimum $w_k$
                    \STATE $\xi \leftarrow \xi - C_{out}$ \label{code:replaceOld}
                \ENDIF
                \STATE $\xi \leftarrow \xi \cup C_{in}$
                \STATE Train all classifiers $C \in \xi - C_{in}$ with $DC$ \label{code:thentrain}
            \ENDIF
        \ENDWHILE
    \end{algorithmic}
\end{algorithm}

\begin{algorithm}
    \caption{GOOWE-ML: Train Optimum Weights}
    \label{alg:goowe-weight}
    \begin{algorithmic}[1]
        \REQUIRE $DC$: one or more data instances, $\xi$: Ensemble of Classifiers, $A$: square matrix, $d$: remainder vector
        \ENSURE The matrix $A$ and the vector $d$, ready for optimum weight assignment 
        \IF{$A$ is null or $d$ is null}
            \STATE Initialize square matrix $A$ of size $K \times K$
            \STATE Initialize remainder vector $d$ of size $K$
        \ENDIF
        \FORALL{instances $x^t \in DC$}
            \STATE $y^t \leftarrow $ true relevance vector of $x^t$ \hfill \COMMENT{To be used in Eqn.\ref{eqnford}}
            \STATE $A \leftarrow A + A^{t}$; \hfill \COMMENT{Eqn.\ref{eqnfora}} \label{code:AeqnLine}
            \STATE $d \leftarrow d + d^{t}$; \hfill \COMMENT{Eqn.\ref{eqnford}} \label{code:deqnLine}
        \ENDFOR
    \end{algorithmic}
\end{algorithm}

\subsection{Complexity Analysis}

Let the prediction of a component classifier in the ensemble take $O(c)$ time. Also, notice that the ensemble size is of order $O(K)$, since ensemble is not fully formed only for the first $K$ chunks and the size is always $K$ afterwards. 

For a data chunk, each component classifier predicts each data instance, which takes $O(h K c)$ time, as the size of each data chunk in the stream are the same and $h$. At the same time, the square matrix $A$ and the remainder vector $d$ are filled using each pair of relevance scores of the components for each label and each instance, which takes $O(h K^2 L)$ time. Then, the linear system $Aw=d$ is solved, where $A$ is of size $K$. Solving this linear system with no complex optimization methods take at most $O(K^3)$ time \cite{bojanczyk1984complexity} (where there are more complex but asymptotically better methods). This loop continues for $N/h$ many chunks. Thus, the whole process has the complexity of:

\begin{equation}
\label{complexity-analysis}
O\Bigg( \frac{N}{h} \Big( (h K c + h K^2 L) + K^3 \Big) \Bigg) = O\Bigg( N \Big( K c + K^2 L + \frac{K^3}{h} \Big) \Bigg)
\end{equation}

Here, the term with $(K c)$, $(K^2 L)$ and $(K^3 /h)$ represents the time complexity of prediction, training and optimal weight assignment respectively.

$c$ is generally small, since most of the models use Hoeffding Trees and its derivatives as their base classifiers. Therefore, the term with $(K^2 L)$ dominates the sum there in the Eqn. \ref{complexity-analysis}. When the terms $(K^2 L)$ and $(K^3 / h)$ are compared, it can be also noticed that the former always dominates the latter: $h$'s magnitude is of hundreds or thousands, whereas $K$'s magnitude is generally of tens. As a result, $L$ (a whole number) will be always higher than $(K/h)$ (a fraction that is $< 1$). As a consequence, the algorithm has overall $O(N K^2 L)$ complexity.

\section{Experimental Design}

\subsection{Datasets}

To understand how densely multi-labeled a dataset is, \textit{Label Cardinality} and \textit{Label Density} are used. Label Cardinality is the average number of relevant labels of the instances in $\mathcal{D}$; Label Density is the Label Cardinality per number of labels \cite{tsoumakas2006overview}, indicating the percentage of labels that are relevant on average.

\vspace{-0.2cm}\begin{align*}{}
  LC(\mathcal{D}) &= \frac{1}{N} \sum_{i=1}^N |y^i| & LD(\mathcal{D}) &= \frac{LC(\mathcal{D})}{L} = \frac{1}{LN} \sum_{i=1}^N |y^i|
\end{align*}

Our experiments are conducted on 7 datasets\footnote{Datasets are downloaded from MEKA's webpage. Available at: \url{https://sourceforge.net/projects/meka/files/Datasets/}.} that are from diverse application domains (genes, newspapers, aviation safety reports and so on), given in Table \ref{datasets}. These datasets are extensively used in the literature \cite{osojnik2017isoup, sousa2018multi, read2012scalable}.

\begin{table}[!ht]
\centering
\caption{Table of Multi-Label Datasets} 
\vspace{-0.2cm}
\label{datasets}
\begin{tabular}{l l@{\hspace{0.9\tabcolsep}} r@{\hspace{0.9\tabcolsep}} r r l l}
 \textbf{Source $\mathcal{D}$} & \textbf{Domain} & \textbf{N} & \textbf{M} &\textbf{L} & \textbf{LC$(\mathcal{D})$} & \textbf{LD$(\mathcal{D}$)}\\
 \hline
 20NG $^{b}$ & Text      & 19,300     & 1,006    & 20     & 1.020 & 0.051\\
 Yeast $^{n}$ & Biology    & 2,417     & 103     & 14    & 4.237 & 0.303\\
 Ohsumed $^{b}$ & Text     & 13,529    & 1,002     & 23    & 1.660 & 0.072\\
 Slashdot $^{b}$ & Text    & 3,782     & 1,079     & 22    & 1.180 & 0.053\\
 Reuters $^{n}$ & Text     & 6,000     & 500      & 101   & 2.880 & 0.028\\
  IMDB $^{b}$ & Text      & 120,919   & 1,001     & 28    & 2.000	& 0.071\\
 TMC2007 $^{b}$ & Text     & 28,596    & 500      & 22    & 2.160 & 0.098\\
 \hline
\end{tabular}
\caption*{The superscripts after the name of the dataset indicates that the features in that dataset is binary ($^{b}$) or numeric ($^{n}$).}
\end{table}

\subsection{Evaluating Multi-label Learners}

Multi-label evaluation metrics that are widely used throughout the studies in the field are divided into two groups \cite{zhang2014review}: (1) \textit{Instance-Based Metrics}, (2) \textit{Label-Based Metrics}. These two metrics indicate how well the algorithms perform. In addition to these, efficiency of the performing algorithms can be measured, which indicates how much resources they consume. Hence, (3) \textit{Efficiency Metrics} is added to the evaluation. In the Tables \ref{experimental-results-1} and \ref{experimental-results-2}, $\uparrow$ ($\downarrow$) next to the metric indicates that the corresponding metric's score is to be maximized (minimized).

\subsubsection{Instance-Based Metrics}

Instance-based metrics are evaluated for every instance and averaged over the whole dataset. Exact Match, Hamming Score, and Instance-Based \{Accuracy, Precision, Recall, F1-Score\} \cite{zhang2014review} are used in this study.





\subsubsection{Label-Based Metrics}

Label-based metrics are evaluated for every label and averaged over examples within each individual label. Macro and Micro-Averaged Precision, Recall and F1 Score \cite{gibaja2014review} are used in this study.
 
 



\subsubsection{Efficiency Metrics}

Finally, to measure the efficiency of the algorithms, the execution time and memory consumption of each algorithm are monitored.

\subsection{Experimental Setup}

Experiments are implemented in MOA \cite{bifet2011moa}, utilizing multi-label methods in MEKA \cite{read2016meka}. The evaluation of each algorithm is \textit{prequential} \cite{gama2009issues}. An incoming data instance is first tested by classifiers (see Alg.\ref{alg:goowe}:\ref{code:test}); evaluation measures corresponding the prediction are recorded, and then, that data instance is used to train classifiers, as well as the updated weighting scheme (see Alg.\ref{alg:goowe}:\ref{code:trainoptimumweights},\ref{code:thentrain}). This is also called \textit{Interleaved-Test-Then-Train (ITTT)} approach and is widely common in algorithms in streaming settings.

If an ensemble is batch-incremental, then the ensemble is trained at the end of each batch (i.e. whenever a data chunk is filled). The evaluation of ensembles are started after the first learner in the ensemble is formed. We used fixed number of 10 classifiers as the ensemble size, mimicking the previously conducted experiments to enable comparison \cite{read2012scalable, osojnik2017isoup}. For incremental evaluation of the classifiers, we used window-based evaluation with the window size $\{100, 250, 500, 1000\}$ according to the size of datasets \footnote{The source code is available at \url{https://github.com/abuyukcakir/gooweml}. The results can be reproduced using the aforementioned datasets. The program outputs the predictive performance measures, time and memory consumption, as well as incremental evaluations for each window.}.

All experiments are conducted on a machine with an Intel Xeon E3-1200 v3 @ 3.40GHz processor and 16GB DDR3 RAM.

We experimented with 4 GOOWE-ML models (referred with their abbreviations from now onward):

\begin{itemize}
    \item \textbf{GOBR:} the components use BR Transformation.
    \item \textbf{GOCC:} the components use CC Transformation.
    \item \textbf{GOPS:} the components use PS Transformation.
    \item \textbf{GORT:} the components use iSOUP Regression Trees.
\end{itemize}

We have 7 baseline models. Four of the baselines use fixed-sized windows with no concept drift detecting mechanism: \textbf{EBR} \cite{read2009classifier}, \textbf{ECC} \cite{read2009classifier}, \textbf{EPS} \cite{read2008pruned}, \textbf{EBRT} \cite{osojnik2017isoup}, whereas 3 of them use ADWIN as their concept drift detector: \textbf{EaBR} \cite{read2012scalable}, \textbf{EaCC} \cite{read2012scalable}, and \textbf{EaPS} \cite{read2012scalable}). In all models, BR and CC transformations use a Hoeffding Tree classifier whereas the PS transformation uses a Naive Bayes classifier.

\subsection{Evaluation of Statistical Significance}

We evaluated the aforementioned algorithms using multi-label example-based, and label-based evaluation metrics, as well as efficiency metrics. To check the statistical significance among the algorithms, we used \textit{Friedman test with Nemenyi post-hoc analysis} \cite{demvsar2006statistical}. We applied the Friedman test with $\alpha = 0.05$ where the null hypothesis is that all of the measurements come from the same distribution. If the null hypothesis is failed, then Nemenyi post-hoc analysis is applied to see which algorithms that performed statistically significantly better than which others.

The result of Friedman-Nemenyi Test can be seen in the \textit{Critical Distance Diagram}s where each algorithm is sorted according to their average ranks for a given metric on a number line, and the algorithms that are within the critical distance of each other (that are not statistically significantly better than each other) are linked with a line. These diagrams compactly show Nemenyi Significance. Better models have lower average rank, and therefore on the right side of a Critical Distance Diagram. The Critical Distance for Nemenyi Significance is calculated as follows \cite{demvsar2006statistical}:

\begin{equation}
    CD = q_{\alpha, m} \sqrt{\frac{m (m+1)}{6 |\mathcal{D}|}}
\end{equation}

where $m$ is the number of models that are being compared, and $|\mathcal{D}|$ number of datasets that are experimented on. Plugging in $m=11$, $q_{\alpha=0.05, m=11} = 3.219$ (from Critical Values Table for Two-Tailed Nemenyi Test \footnote{Available at: \url{http://www.cin.ufpe.br/~fatc/AM/Nemenyi_critval.pdf} }) and $|\mathcal{D}|=7$, we get $CD=5.707$ as our Critical Distance.

\section{Results and Discussion}

\subsection{Predictive Performance}

\begin{table*}[]
\centering
\caption{Experimental Results: Example and Label-based Metrics}
\vspace{-0.3cm}
\label{experimental-results-1}         
\scalebox{0.72}{
\begin{tabular}{llllllllcllllllllc}
\multicolumn{1}{l}{} & 20NG           & Yeast          & Ohsumed        & Slashdot       & Reuters        & IMDB           & TMC7           & \multicolumn{1}{l}{}                    & \multicolumn{1}{l}{} & 20NG           & Yeast          & Ohsumed        & Slashdot       & Reuters        & IMDB           & TMC7           & \multicolumn{1}{l}{} \\ \hline
\rowcolor{gray!40} \multicolumn{8}{l}{\textbf{(a) Example-Based F1 Score ($F1_{ex}$)  $\uparrow$}}                                                             & \multicolumn{1}{c|}{\textbf{Avg. Rank}} & \multicolumn{8}{l}{\textbf{(b) Micro-Averaged F1 Score ($F1_{micro}$) $\uparrow$}}                                                          & \textbf{Avg. Rank}   \\
\textbf{GOBR}        & 0.364          & 0.650          & 0.307          & 0.189          & 0.076          & 0.283          & 0.623          & \multicolumn{1}{c|}{4.00}               & \textbf{GOBR}        & 0.237          & 0.638          & 0.291          & 0.187          & 0.076          & 0.276          & 0.584          & 4.86                 \\
\textbf{GOCC}        & \textbf{0.442} & \textbf{0.652} & \textbf{0.352} & 0.028          & 0.145          & 0.221          & \textbf{0.668} & \multicolumn{1}{c|}{{\ul 2.57}}         & \textbf{GOCC}        & \textbf{0.516} & \textbf{0.640} & \textbf{0.410} & 0.050          & 0.196          & 0.228          & 0.634          & {\ul 2.71}           \\
\textbf{GOPS}        & 0.224          & 0.644          & 0.331          & \textbf{0.405} & \textbf{0.252} & \textbf{0.333} & 0.485          & \multicolumn{1}{c|}{3.00}               & \textbf{GOPS}        & 0.206          & 0.629          & 0.298          & \textbf{0.315} & \textbf{0.210} & \textbf{0.314} & 0.447          & 3.43                 \\
\textbf{GORT}       & 0.196          & 0.607          & 0.297          & 0.189          & 0.078          & 0.283          & 0.452          & \multicolumn{1}{c|}{5.71}               & \textbf{GORT}       & 0.153          & 0.598          & 0.270          & 0.187          & 0.077          & 0.277          & 0.439          & 6.57                 \\ \hline
\textbf{EBR}         & 0.365          & 0.638          & 0.23           & 0.023          & 0.106          & 0.075          & 0.654          & \multicolumn{1}{c|}{4.71}               & \textbf{EBR}         & 0.499          & 0.631          & 0.294          & 0.041          & 0.141          & 0.099          & 0.638          & 4.29                 \\
\textbf{ECC}         & 0.349          & 0.632          & 0.217          & 0.020          & 0.098          & 0.016          & 0.643          & \multicolumn{1}{c|}{6.43}               & \textbf{ECC}         & 0.486          & 0.625          & 0.280          & 0.037          & 0.134          & 0.025          & 0.631          & 6.14                 \\
\textbf{EPS}         & 0.096          & 0.584          & 0.213          & 0.269          & 0.148          & 0.133          & 0.330          & \multicolumn{1}{c|}{6.71}               & \textbf{EPS}         & 0.115          & 0.584          & 0.216          & 0.286          & 0.162          & 0.138          & 0.342          & 7.00                 \\
\textbf{EBRT}        & 0.100          & 0.509          & 0.056          & 0.001          & 0.000          & 0.001          & 0.008          & \multicolumn{1}{c|}{10.57}              & \textbf{EBRT}        & 0.174          & 0.519          & 0.076          & 0.001          & 0.000          & 0.001          & 0.008          & 10.58                \\
\textbf{EaBR}        & 0.341          & 0.638          & 0.202          & 0.018          & 0.059          & 0.031          & 0.661          & \multicolumn{1}{c|}{6.57}               & \textbf{EaBR}        & 0.477          & 0.632          & 0.266          & 0.033          & 0.081          & 0.041          & \textbf{0.640} & 5.71                 \\
\textbf{EaCC}        & 0.156          & 0.633          & 0.005          & 0.020          & 0.004          & 0.001          & 0.646          & \multicolumn{1}{c|}{8.14}               & \textbf{EaCC}        & 0.262          & 0.627          & 0.007          & 0.037          & 0.007          & 0.002          & 0.632          & 7.71                 \\
\textbf{EaPS}        & 0.109          & 0.578          & 0.200          & 0.258          & 0.183          & 0.104          & 0.384          & \multicolumn{1}{c|}{6.85}               & \textbf{EaPS}        & 0.180          & 0.580          & 0.205          & 0.278          & 0.200          & 0.118          & 0.378          & 6.71                 \\ \hline \hline
\rowcolor{gray!40} \multicolumn{8}{l}{\textbf{(c) Hamming Score $\uparrow$}}                                                                                   & \multicolumn{1}{c|}{\textbf{Avg. Rank}} & \multicolumn{8}{l}{\textbf{(d) Example-Based Accuracy ($Acc_{ex}$) $\uparrow$}}                                                             & \textbf{Avg. Rank}   \\
\textbf{GOBR}        & 0.749          & 0.769          & 0.738          & 0.625          & 0.707          & 0.727          & 0.886          & \multicolumn{1}{c|}{9.86}               & \textbf{GOBR}        & 0.239          & 0.508          & 0.184          & 0.106          & 0.040          & 0.164          & 0.457          & 4.57                 \\
\textbf{GOCC}        & 0.952          & 0.771          & 0.932          & 0.946          & 0.984          & 0.887          & 0.916          & \multicolumn{1}{c|}{5.57}               & \textbf{GOCC}        & \textbf{0.391} & \textbf{0.509} & \textbf{0.277} & 0.025          & 0.120          & 0.138          & 0.515          & {\ul 3.00}           \\
\textbf{GOPS}        & 0.769          & 0.754          & 0.830          & 0.872          & 0.956          & 0.836          & 0.854          & \multicolumn{1}{c|}{9.29}               & \textbf{GOPS}        & 0.137          & 0.504          & 0.211          & \textbf{0.299} & 0.160          & \textbf{0.204} & 0.327          & 3.29                 \\
\textbf{GORT}       & 0.624          & 0.716          & 0.730          & 0.644          & 0.720          & 0.732          & 0.815          & \multicolumn{1}{c|}{10.57}              & \textbf{GORT}       & 0.115          & 0.454          & 0.178          & 0.107          & 0.040          & 0.164          & 0.298          & 6.71                 \\ \hline
\textbf{EBR}         & \textbf{0.961} & 0.786          & \textbf{0.936} & 0.946          & \textbf{0.986} & 0.925          & 0.934          & \multicolumn{1}{c|}{2.14}               & \textbf{EBR}         & 0.352          & 0.502          & 0.191          & 0.020          & 0.098          & 0.055          & 0.520          & 4.29                 \\
\textbf{ECC}         & \textbf{0.961} & 0.786          & \textbf{0.936} & \textbf{0.947} & \textbf{0.986} & 0.928          & 0.934          & \multicolumn{1}{c|}{{\ul 1.57}}         & \textbf{ECC}         & 0.337          & 0.493          & 0.180          & 0.018          & 0.093          & 0.012          & 0.511          & 6.14                 \\
\textbf{EPS}         & 0.924          & 0.764          & 0.918          & 0.937          & 0.985          & 0.919          & 0.911          & \multicolumn{1}{c|}{7.29}               & \textbf{EPS}         & 0.094          & 0.460          & 0.180          & 0.260          & 0.143          & 0.105          & 0.246          & 6.29                 \\
\textbf{EBRT}        & 0.952          & 0.773          & 0.930          & 0.946          & \textbf{0.986} & \textbf{0.929} & 0.902          & \multicolumn{1}{c|}{4.00}               & \textbf{EBRT}        & 0.100          & 0.372          & 0.049          & 0.001          & 0.000          & 0.001          & 0.007          & 10.57                \\
\textbf{EaBR}        & \textbf{0.961} & 0.786          & 0.935          & 0.946          & \textbf{0.986} & 0.928          & \textbf{0.935} & \multicolumn{1}{c|}{2.00}               & \textbf{EaBR}        & 0.330          & 0.502          & 0.169          & 0.016          & 0.056          & 0.024          & \textbf{0.529} & 6.14                 \\
\textbf{EaCC}        & 0.955          & \textbf{0.787} & 0.928          & \textbf{0.947} & \textbf{0.986} & \textbf{0.929} & 0.934          & \multicolumn{1}{c|}{2.29}               & \textbf{EaCC}        & 0.152          & 0.495          & 0.004          & 0.018          & 0.004          & 0.001          & 0.516          & 7.71                 \\
\textbf{EaPS}        & 0.950          & 0.767          & 0.918          & 0.937          & 0.985          & 0.924          & 0.913          & \multicolumn{1}{c|}{6.71}               & \textbf{EaPS}        & 0.108          & 0.455          & 0.170          & 0.250          & \textbf{0.179} & 0.083          & 0.290          & 6.43                 \\ \hline
\end{tabular}
}   
\end{table*}

\begin{figure*}[!ht]
\includegraphics[width=1\linewidth]{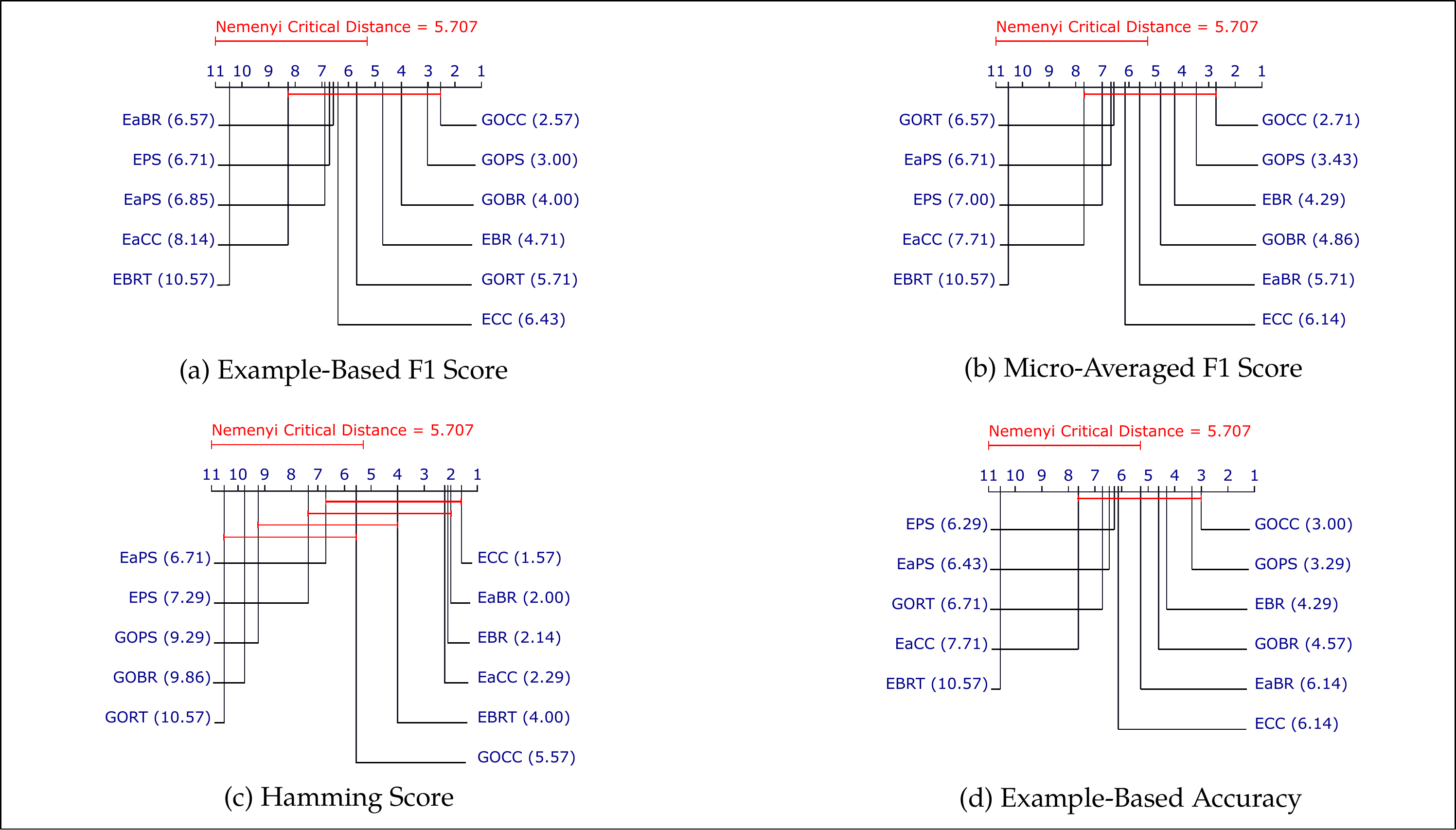}
\vspace{-0.6cm} \caption{Critical Distance Diagrams for the Predictive Performance Metrics (given in Table \ref{experimental-results-1}).}
\label{cdd-diagrams-1}
\end{figure*}

Example-Based F1 Score, Micro-Averaged F1 Score, Hamming Score and Example-Based Accuracy are given in Table \ref{experimental-results-1} for each model on each dataset. Winner models of each dataset for the metrics are shown in bold in the table. Precision and Recall scores are omitted, since we report the F1 Scores in Table \ref{experimental-results-1}, which is calculated as the harmonic mean of the two. Exact Match scores are omitted, since it is a very strict metric and the scores tend to be near zero for each algorithm especially when $|\mathcal{L}|$ is large. 

Before starting to analyze models individually, let us look at the big picture: It is apparent that the predictive performance of a streaming multi-label model highly depends on the dataset. Looking at the results, no single model is clearly better than the rest, regardless of the dataset that it has run on. For instance, PS transformation-based ensembles (GOPS, EPS and EaPS) did relatively better in the Slashdot, Reuters and IMDB datasets; whereas the ensembles with BR and CC transformations were clearly superior in the Yeast, Ohsumed and TMC2007 datasets. 

It can be observed in the Table \ref{experimental-results-1} and Figure \ref{cdd-diagrams-1}, GOOWE-ML-based classifiers performed better than Online Bagging and ADWIN Bagging models consistently over all datasets. Especially \textbf{GOCC} and \textbf{GOPS} placed 1st and 2nd respectively, in every performance metric, except Hamming Score. More detailed discussion on the Hamming Score and its relation to the Precision and Recall scores of the models are provided below in a separate section.

Read et al. \cite{read2012scalable} previously claimed that instance-incremental methods are better than batch-incremental methods in the MLSC task. However, our experimental evidence shows that our batch-incremental ensemble performs better than the state-of-the-art instance-incremental models in almost every performance metric (again, except Hamming Score).

\subsection{Efficiency}

\begin{figure*}[!ht]
\includegraphics[width=1\linewidth]{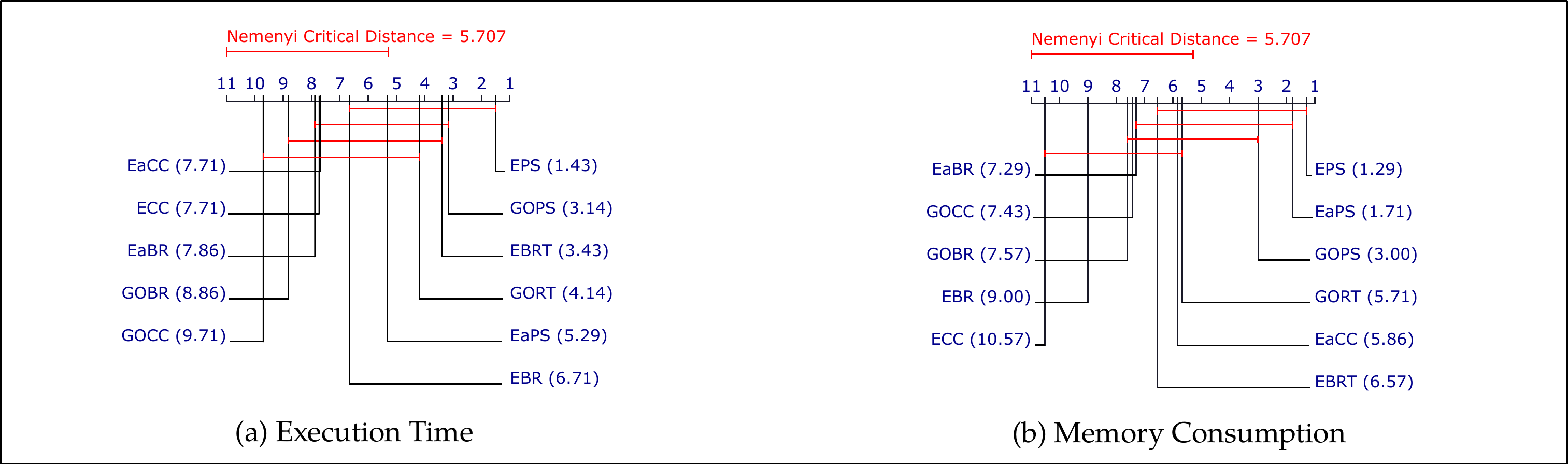}
\vspace{-0.5cm} \caption{Critical Distance Diagrams for the Efficiency Metrics (given in Table \ref{experimental-results-2}).}
\label{cdd-diagrams-2}
\end{figure*}

Results for the Execution Time and Memory Consumption of the models, and the corresponding Critical Distance Diagrams are given in Table \ref{experimental-results-2} and Figure \ref{cdd-diagrams-2}, respectively. It is clear that time and memory efficiency of an MLSC ensemble is highly correlated with the problem transformation method that its component classifiers use. Ensembles that used PS Transformation (GOPS, EPS, EaPS) are ranked consistently higher in terms of both time and memory efficiency. Indeed, as it can be seen in Figure \ref{cdd-diagrams-2}, EPS and GOPS are among the top 3 for both of the metrics.

Models with iSOUP Tree are among the fastest, but their memory consumption is significantly high compared to the PS-based ensembles. Considering GORT and EBRT's relatively underwhelming predictive performance (see Figure \ref{experimental-results-1}), PS-based ensembles should be preferable over ensembles of iSOUP Regression Trees.

BR and CC Transformation-based models performed similarly within datasets across ensembling techniques. Their execution time and memory consumption is nearly identical with a few exceptions (where ADWIN Bagging models had significantly lower memory consumption due to resetting component classifiers many times). Having similar resource consumptions, GOCC can be preferred due to its greater predictive performance.

\subsection{On Hamming Scores in Datasets with Large Labelsets}

Consider the prediction and the ground truth vector of a given data instance. Let $TP$, $FP$, $FN$ and $TN$ denote the number of true positives, false positives, false negatives and true negatives. For instance, $FP$ is the number of labels that are predicted as relevant but are not. Then, Hamming Score for that instance can be calculated as $ \frac{TP+TN}{TP+FP+FN+TN} $.

For a multi-label dataset with a fairly large labelset and low label density, $TN$ in the numerator and the denominator will dominate this score and Hamming Score will yield mis-interpretable results. Take IMDB dataset ($|\mathcal{L}| = 28$, $LD(\mathcal{D}) = 0.071$) for example: \textbf{GOPS} was the clear winner in terms of $F1_{ex}$, $Acc_{ex}$ and $F1_{micro}$, yielding $20\%$ better scores than its closest competitor (which was \textbf{GOBR}). Despite performing this well, GOPS had a considerably \textit{low} Hamming Score (\textbf{0.836}) with respect to the Online Bagging-based models (all of them around \textbf{0.928}). Here, one can argue that perhaps the Hamming Score is the true indicator of success in MLL, and therefore Online Bagging-based models performed better. However, this hypothesis cannot be correct, since even a dummy classifier that predicts every single label as irrelevant (0) yields Hamming Score of \textbf{0.929}! Additionally, the reason why GOOWE-ML-based models have smaller Hamming Scores is that they have high $FP$ (hence low $TN$) values in the contingency table. In other words, GOOWE-ML-based models are \textit{Low Precision - High Recall} models. They eagerly predict labels as relevant. On the other hand, Online Bagging-based models are \textit{High Precision - Low Recall} models. They predict few labels as relevant at each data instance. As a consequence, they are more confident about their predictions, but they miss many relevant labels due to being more conservative. This dichotomy is shown for 3 datasets with low label densities in Table \ref{precision-vs-recall}, where the higher value among Precision and Recall is shown in bold for each model and dataset.

\begin{table*}[]
\centering
\vspace{-0.2cm}
\caption{Experimental Results: Efficiency Metrics}
\vspace{-0.4cm}
\label{experimental-results-2}         
\scalebox{0.68}{
\begin{tabular}{lrrrrrrrclrrrrrrrc}
\multicolumn{1}{l}{} & 20NG  & Yeast & Ohsumed & Slashdot & Reuters & IMDB        & TMC7  & \multicolumn{1}{r}{}                    &                & 20NG     & Yeast  & Ohsumed  & Slashdot & Reuters  & IMDB          & TMC7     & \multicolumn{1}{l}{} \\ \hline
\rowcolor{gray!40} \multicolumn{8}{l}{\textbf{(a) Execution Time (seconds) $\downarrow$}}                    & \multicolumn{1}{c|}{\textbf{Avg. Rank}} & \multicolumn{8}{l}{\textbf{(b) Memory Consumption (MB) $\downarrow$}}                          & \textbf{Avg. Rank}   \\
\textbf{GOBR}        & 2,631 & 28    & 2,310    & 537      & 2,366   & 31,769      & 1,942  & \multicolumn{1}{c|}{8.86}               & \textbf{GOBR}  & 1,685.82 & 18.40  & 1,364.32 & 381.98   & 1,029.23 & 4,384.09      & 780.58   & 7.57                 \\
\textbf{GOCC}        & 2,591 & 33    & 2,314    & 544      & 2,555   & 34,348      & 1,990  & \multicolumn{1}{c|}{9.71}               & \textbf{GOCC}  & 1,429.26 & 24.85  & 1,229.32 & 351.74   & 1,261.34 & 6,284.62      & 748.33   & 7.43                 \\
\textbf{GOPS}        & 670   & 8     & 522     & 129      & 115     & 5,098       & 181   & \multicolumn{1}{c|}{3.14}               & \textbf{GOPS}  & 76.30    & 2.03   & 43.55    & 29.38    & 15.57    & 75.68         & 41.88    & 3.00                 \\
\textbf{GORT}       & 390   & 47    & 435     & 68       & 412     & 3,719       & 333   & \multicolumn{1}{c|}{4.14}               & \textbf{GORT} & 431.76   & 198.81 & 656.16   & 77.92    & 660.38   & 542.24        & 227.70   & 5.71                 \\ \hline
\textbf{EBR}         & 2,246 & 25    & 1,934   & 488      & 1,917   & \textbf{(*)} 101,243 & 1,769 & \multicolumn{1}{c|}{6.71}               & \textbf{EBR}   & 2,152.97 & 17.56  & 1,775.72 & 545.72   & 1,425.88 & \textbf{(*)} 22,119.42 & 1,289.87 & 9.00                 \\
\textbf{ECC}         & 2,270 & 29    & 1,958   & 495      & 2,057   & \textbf{(*)} 48,325  & 1,789 & \multicolumn{1}{c|}{7.71}               & \textbf{ECC}   & 2,171.09 & 28.99  & 1,792.66 & 549.33   & 1,539.00     & \textbf{(*)} 22,380.89 & 1,305.07 & 10.57                \\
\textbf{EPS}         & 383   & 5     & 299     & 99       & 46      & 2,168       & 109   & \multicolumn{1}{c|}{{\ul 1.43}}         & \textbf{EPS}   & 8.40     & 0.97   & 8.53     & 10.38    & 3.67     & 7.55          & 6.34     & {\ul 1.29}           \\
\textbf{EBRT}        & 338   & 63    & 404     & 64       & 389     & 3,919       & 264   & \multicolumn{1}{c|}{3.43}               & \textbf{EBRT}  & 521.96   & 274.61 & 809.76   & 76.70    & 943.59   & 1,826.87      & 234.06   & 6.57                 \\
\textbf{EaBR}        & 2,376 & 35    & 1,997   & 488      & 1,968   & 20,675      & 2,220 & \multicolumn{1}{c|}{7.86}               & \textbf{EaBR}  & 1,997.59 & 17.57  & 1,678.32 & 399.99   & 1,330.31 & 3,522.14      & 93.60    & 7.29                 \\
\textbf{EaCC}        & 2,041 & 40    & 1,622   & 503      & 2,062   & 17,148      & 2,292 & \multicolumn{1}{c|}{7.71}               & \textbf{EaCC}  & 373.03   & 26.39  & 295.09   & 549.35   & 652.92   & 661.76        & 135.09   & 5.86                 \\
\textbf{EaPS}        & 2,393 & 24    & 1,862   & 363      & 402     & 14,361      & 574   & \multicolumn{1}{c|}{5.29}               & \textbf{EaPS}  & 6.25     & 1.50   & 15.80    & 12.15    & 8.05     & 13.53         & 2.56     & 1.71                 \\ \hline
\end{tabular}
}   
\raggedright \small Note. Measurements that are marked with \textbf{(*)} are conducted on a machine with 256 GB RAM. Consistency among the rankings is preserved, as the obtained results do not change the rankings of the efficiency metrics for the IMDB dataset. 

\end{table*}

Observing these, we claim that \emph{in datasets with large labelset and low label density, High Recall models may have considerably low Hamming Scores due to the nature of the metric. Hence, Hamming Score may not be the true indicator of the predictive performance while evaluating multi-label models}.

\begin{table}[!h]
\centering
\caption{Micro Precision (Prec) vs Recall (Rec), and Their Effect on Hamming Score (HS)}
\label{precision-vs-recall}
\scalebox{0.8}{
\vspace{-0.5cm}\begin{tabular}{rccccccccc}
\multicolumn{1}{l}{} & \multicolumn{3}{c}{\textbf{20NG}}                                  & \multicolumn{3}{c}{\textbf{Ohsumed}}                               & \multicolumn{3}{c}{\textbf{Reuters}}          \\ \cline{2-10} 
\multicolumn{1}{l}{} & Prec           & Rec            & \multicolumn{1}{c|}{HS}          & Prec           & Rec            & \multicolumn{1}{c|}{HS}          & Prec           & Rec            & HS          \\ \cline{2-10} 
GOBR                 & 0.140          & \textbf{0.757} & \multicolumn{1}{c|}{0.749}       & 0.181          & \textbf{0.743} & \multicolumn{1}{c|}{0.738}       & 0.040          & \textbf{0.848} & 0.707       \\
GOPS                 & 0.125          & \textbf{0.580} & \multicolumn{1}{c|}{0.769}       & 0.212          & \textbf{0.500} & \multicolumn{1}{c|}{0.830}       & 0.140          & \textbf{0.418} & 0.956       \\ \cline{2-10} 
EBR                  & \textbf{0.753} & 0.373          & \multicolumn{1}{c|}{{\ul 0.961}} & \textbf{0.713} & 0.185          & \multicolumn{1}{c|}{{\ul 0.936}} & \textbf{0.510} & 0.082          & {\ul 0.986} \\
EPS                  & \textbf{0.142} & 0.096          & \multicolumn{1}{c|}{{\ul 0.924}} & \textbf{0.348} & 0.157          & \multicolumn{1}{c|}{{\ul 0.918}} & \textbf{0.361} & 0.105          & {\ul 0.985}
\end{tabular}
}
\begin{tablenotes}
      \small
      \item 
      \item Two GOOWE-ML models and two Online Bagging Models with different problem transformation types are picked. Higher Precision and lower Recall resulted in better Hamming Scores consistently.
    \end{tablenotes}
\end{table}

This hypothesis helps explaining why GOBR and GOCC performed poorly in terms of Hamming Scores in the datasets with relatively lower label densities (such as Slashdot, Reuters and IMDB datasets) whereas both of them were clear winners in the overall predictive performance.

\subsection{Window-Based Evaluation}

\begin{figure*}[!h]
        \centering
        \scalebox{0.69}{
        \begin{tikzpicture}
            \begin{groupplot}[
                group style={
                    group name=my plots,
                    group size=2 by 1,
                    xlabels at=edge bottom,
                    ylabels at=edge left,
                    horizontal sep=2cm,vertical sep=3cm,
                },
                legend columns=-1,
                xtick={0,4,8,12,16,20,24},
                xmin=0, xmax=24,
                ytick={0,0.1,0.2,0.3,0.4,0.5,0.6,0.7},
                ymin=0, ymax=0.7,
                width=0.5\linewidth
                ]
            
            \nextgroupplot[xlabel={Window Number [$\times$ 250] Instances},
            ylabel={Example-Based F1 Score}, 
            title={Reuters Dataset},
            xtick={0,4,8,12,16,20,24},
            xmin=0, xmax=24,
            ytick={0,0.05,0.1,0.15,0.2,0.25,0.3,0.35},
            ymin=0, ymax=0.35,
            legend style={at={(0.97,0.03)}, anchor=south east},
            tick label style={/pgf/number format/fixed}]
         
            \addplot[
                color=blue,
                mark=square*,
            ]
            coordinates {
            (1,0.1047428571)
            (2,0.1932571429)
            (3,0.2672)
            (4,0.29927619047619)
            (5,0.269107936507936)
            (6,0.243466666666666)
            (7,0.262663492063492)
            (8,0.232368253968254)
            (9,0.240816738816739)
            (10,0.264470707070707)
            (11,0.24419393939394)
            (12,0.223766233766234)
            (13,0.222508513708515)
            (14,0.248207215007216)
            (15,0.236025974025975)
            (16,0.237172005772007)
            (17,0.225352525252526)
            (18,0.261378066378067)
            (19,0.231397402597403)
            (20,0.214133910533911)
            (21,0.256396825396826)
            (22,0.228102453102454)
            (23,0.240)
            };
            \addlegendentry{GOPS}
            
            \addplot[
                color=red,
                mark=triangle*,
            ]
            coordinates {
            (0, 0.056)
            (1,0.1729333333)
            (2,0.167333333333333)
            (3,0.194666666666667)
            (4,0.182266666666667)
            (5,0.162)
            (6,0.142)
            (7,0.1456)
            (8,0.13)
            (9,0.138666666666667)
            (10,0.148)
            (11,0.137333333333333)
            (12,0.130666666666667)
            (13,0.134666666666667)
            (14,0.146266666666667)
            (15,0.144)
            (16,0.144666666666667)
            (17,0.152)
            (18,0.202666666666667)
            (19,0.141333333333333)
            (20,0.135333333333333)
            (21,0.145333333333333)
            (22,0.124)
            (23,0.131333333333333)
            };
            \addlegendentry{EPS}
            
            \addplot[
                color=green!60!black,
                mark=otimes*,
            ]
            coordinates {
            (0, 0.008)
            (1,0.111333333333333)
            (2,0.200666666666667)
            (3,0.226)
            (4,0.202933333333333)
            (5,0.158)
            (6,0.179333333333333)
            (7,0.186266666666667)
            (8,0.196666666666667)
            (9,0.188)
            (10,0.206666666666667)
            (11,0.186)
            (12,0.18)
            (13,0.2)
            (14,0.2036)
            (15,0.172666666666667)
            (16,0.2)
            (17,0.209333333333333)
            (18,0.232)
            (19,0.225333333333333)
            (20,0.196)
            (21,0.161333333333333)
            (22,0.1816)
            (23,0.150666666666667)
            };
            \addlegendentry{EaPS}
            
            \nextgroupplot[xlabel={Window Number [$\times$ 1000 instances]},
            ylabel={Example-Based F1 Score}, 
            title={20NG Dataset},
            xtick={0,4,8,12,16,20},
            xmin=0, xmax=19,
            ytick={0,0.1,0.2,0.3,0.4,0.5,0.6,0.7},
            ymin=0, ymax=0.6,
            legend style={at={(0.97,0.03)}, anchor=south east}]

            \addplot[
                color=blue!60!black,
                mark=square*,
            ]
            coordinates {
            (1,0.0196666666666667)
            (2,0.064)
            (3,0.115333333333333)
            (4,0.358466666666667)
            (5,0.434066666666668)
            (6,0.461433333333335)
            (7,0.508400000000002)
            (8,0.464633333333336)
            (9,0.519466666666669)
            (10,0.495233333333336)
            (11,0.50366666666667)
            (12,0.533100000000003)
            (13,0.518900000000002)
            (14,0.554066666666668)
            (15,0.514633333333337)
            (16,0.483852380952383)
            (17,0.516433333333335)
            (18,0.549066666666669)
            };
            \addlegendentry{GOCC}
            
            \addplot[
                color=red!80!black,
                mark=triangle*,
            ]
            coordinates {
            (0, 0.0327182186234818)
            (1,0.002)
            (2,0.046)
            (3,0.296033333333333)
            (4,0.349333333333333)
            (5,0.366066666666667)
            (6,0.382666666666667)
            (7,0.444333333333334)
            (8,0.412166666666667)
            (9,0.4105)
            (10,0.406500000000001)
            (11,0.420666666666667)
            (12,0.4325)
            (13,0.445666666666667)
            (14,0.492333333333334)
            (15,0.455833333333334)
            (16,0.447166666666667)
            (17,0.445)
            (18,0.518000000000001)
            };
            \addlegendentry{EBR}
            
            \addplot[
                color=green!40!black,
                mark=otimes*,
            ]
            coordinates {
            (0, 0.0527182186234818)
            (1,0.0)
            (2,0.021)
            (3,0.117666666666667)
            (4,0.254)
            (5,0.288)
            (6,0.36)
            (7,0.397833333333333)
            (8,0.384666666666667)
            (9,0.3935)
            (10,0.383666666666667)
            (11,0.404)
            (12,0.438833333333334)
            (13,0.467166666666667)
            (14,0.4885)
            (15,0.453333333333334)
            (16,0.440333333333333)
            (17,0.4505)
            (18,0.547833333333334)
            };
            \addlegendentry{EaBR}
            
        \end{groupplot}
    \end{tikzpicture}
        }
    \vspace{-0.3cm} \caption{Window-Based Evaluation of Models: Example-Based F1 Score for Reuters and 20NG datasets.}
    \label{window-based-evaluation}
\end{figure*}
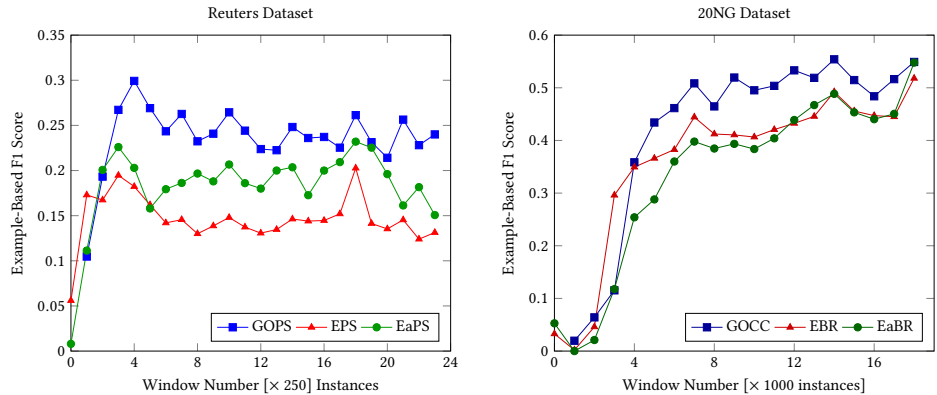

Figure \ref{window-based-evaluation} presents window-based evaluation for two datasets and three models, where the sliding window size is equal to the chunk size, i.e. $n=h$. To this end, we evaluate each window's performance using $F1_{ex}$ measurement. For each group of models, the best performing strategy is chosen for the given dataset---e.g. in Reuters dataset, GOPS, EPS and EaPS performed the best among GOOWE-ML, Online Bagging and ADWIN Bagging models, respectively.

It can be seen that GOOWE-ML-based models do not predict in the first evaluation window, since no training has been done while waiting the first chunk to be filled. In both of the datasets, we observe the optimal weight assignment strategy in effect: after the first few chunks, GOOWE-ML-based model's predictions continually yield better performance than its competitors.

\section{Conclusion}

We present an online batch-incremental multi-label stacked ensemble, GOOWE-ML, that constructs a spatial model for the relevance scores of its classifiers, and uses this model to assign optimal weights to its component classifiers. Our experiments show that GOOWE-ML models outperform the most prominent Online Bagging and ADWIN Bagging models. Two of the GOOWE-ML-based ensembles especially stand out: \textbf{GOCC} is the clear winner in terms of overall predictive performance, ranking first in $Acc_{ex}$, $F1_{ex}$ and $F1_{micro}$ scores. \textbf{GOPS}, on the other hand, is the best compromise between predictive performance and resource consumption among all models, yielding strong performance with very conservative time and memory requirements. In addition, we argue that Hamming Score can be deceptively low for models with low Precision and high Recall and support this claim by experimental evidence.

In the future, we plan to investigate optimal ensemble size for MLSC in relation to the dimensionality of the feature set, number of labels, and label cardinality and density. We also plan to study the performance of GOOWE-ML for concept-evolving multi-label data streams, in which the labelset can be updated with new labels.

\bibliographystyle{ACM-Reference-Format}


\end{document}